\documentclass{article} 
\usepackage[dvipsnames]{xcolor}         
\usepackage{iclr2023_conference,times}
\usepackage{hyperref}
\usepackage{url}
\usepackage{booktabs}       
\usepackage{amsfonts}       
\usepackage{nicefrac}       
\usepackage{microtype}      
\usepackage{graphicx}
\usepackage{wrapfig}
\usepackage{colortbl}
\usepackage{amsmath}
\usepackage{placeins}
\usepackage{adjustbox}
\usepackage{ifthen}
\usepackage{etoolbox}

\title{Contrastive Alignment of Vision to Language Through Parameter-Efficient Transfer Learning
}

\author{Zaid Khan, Yun Fu  \\
Northeastern University, Boston, USA \\
\texttt{\{khan.za, y.fu\}@northeastern.edu}
}

\newcommand{\ice}{\fcolorbox{black}{CornflowerBlue}{\rule{0pt}{1pt}\rule{1pt}{0pt}}}

\newcommand{\fire}{\fcolorbox{black}{Orange}{\rule{0pt}{1pt}\rule{1pt}{0pt}}}

\newcommand{\emoji}[1]{\ifstrequal{#1}{ice}{\ice}{\fire}}

\iclrfinalcopy 
\begin{document}

\maketitle

\begin{abstract}
Contrastive vision-language models (e.g. CLIP) are typically created by updating all the parameters of a vision model and language model through contrastive training. Can such models be created by a small number of parameter updates to an already-trained language model and vision model? The literature describes techniques that can create vision-language models by updating a small number of parameters in a language model, but these require already aligned visual representations and are non-contrastive, hence unusable for latency-sensitive applications such as neural search. We explore the feasibility and benefits of parameter-efficient contrastive vision-language alignment through transfer learning: creating a model such as CLIP by minimally updating an already-trained vision and language model. We find that a minimal set of parameter updates ($<$7\%) can achieve the same performance as full-model training, and updating specific components ($<$1\% of parameters) can match 75\% of full-model training. We describe a series of experiments: we show that existing knowledge is conserved more strongly in parameter-efficient training and that parameter-efficient scaling scales with model and dataset size. Where paired-image text data is scarce but strong multilingual language models exist (e.g. low resource languages), parameter-efficient training is even preferable to full-model training. Given a fixed compute budget, parameter-efficient training allows training larger models on the same hardware, achieving equivalent performance in less time. Parameter-efficient training hence constitutes an energy-efficient and effective training strategy for contrastive vision-language models that may be preferable to the full-model training paradigm for common use cases.
Code and weights at \url{https://github.com/codezakh/LilT}.
\end{abstract}

\section{Introduction}
Advances in transfer learning within the field of natural language processing \citep{adapters,bitfit} have shown that when adapting to a novel task, updates to a small percentage of neurons ($<1$\%) in large, pretrained transformer-based language models can achieve nearly equivalent results to finetuning the entire model.
\cite{Sung2021VLAdapterPT} showed that given the existence of already-aligned visual representations (e.g. CLIP's visual encoder) only a small number ($4$\%) of parameters in a pretrained language model need to be updated for the language model to complete tasks such as visual question answering using the already-aligned visual representations.
However, \textit{the creation of aligned vision and language representations typically involves updating all the parameters} of a language model and a vision model, often randomly initialized \citep{clip}.
\cite{lit} find that if the weights of a pretrained vision model are used as an initialization, only the neurons of the language model need to be updated to align the visual and language representations and match or exceed the performance of full-model training, resulting in a $50$\% reduction in trainable parameters.
We take this line of investigation to its natural conclusion, asking --- \textit{given that strong, pretrained vision and language models both exist, can we minimally update both of their parameters to align their representations?}

Answering this question is valuable for two reasons.
From a practical perspective, contrastive vision-language alignment constitutes a form of large-scale pretraining and hence a heavy energy expenditure. 
Methods for parameter-efficient transfer learning result in significantly reduced GPU memory requirements, and can therefore lower energy costs.
Second, collecting millions of images with textual annotations is prohibitively expensive when millions of image-text pairs cannot be scraped from the internet, such as in the case of low resource languages or images from domains that require expert descriptions. 
In these cases, transfer learning by maximally preserving knowledge from strong, unimodal pretraining becomes compelling. 
Our contributions can be summarized as follows.
\begin{itemize}
    \item We show contrastive vision-language models can be created by updates to a relatively small ($<$7\%) set of parameters in pretrained vision and language models, which we dub \textbf{LilT} (Locked image-language tuning) for brevity.
    \item We conduct an detailed empirical study of combinations and interactions of various methods for parameter-efficient transfer learning.
    \item We show that contrastive vision-language models created with parameter-efficient transfer learning conserve useful existing knowledge from their initializations better than full model finetuning, and this has benefits in realistic scenarios.
\end{itemize}

\textbf{Limitations} Similar to \cite{virtex}, we conduct most of our experiments on the COCO dataset, and conduct additional scaling experiments with a larger dataset of $1.5$M pairs. 
There is a possibility that our conclusions may not hold beyond this range.
Second, we choose to focus on zero-shot classification and information retrieval tasks. 
Our conclusions may not hold for other uses of image-text embeddings, such as using them as input for downstream vision-language tasks.
Finally, we explicitly limit the scope of the study to transformer-based contrastive vision-language models.
Thus, our conclusions may not apply to those based on other architectures.
Despite these limitations, we believe our conclusions are useful because there are realistic situations in which there are much fewer than $1.5$M image-text pairs (e.g. low resource languages) available.

\textbf{Outline} First, we cover background material (\S\ref{sec:background}), then introduce our approach of parameter-efficient transfer learning for contrastive vision-language alignment (\S\ref{sec:methods}).
We then describe experiments and a discussion of experimental results (\S\ref{sec:experiments}), followed by related work (\S\ref{sec:related}).

\begin{figure}
  \centering
  \includegraphics{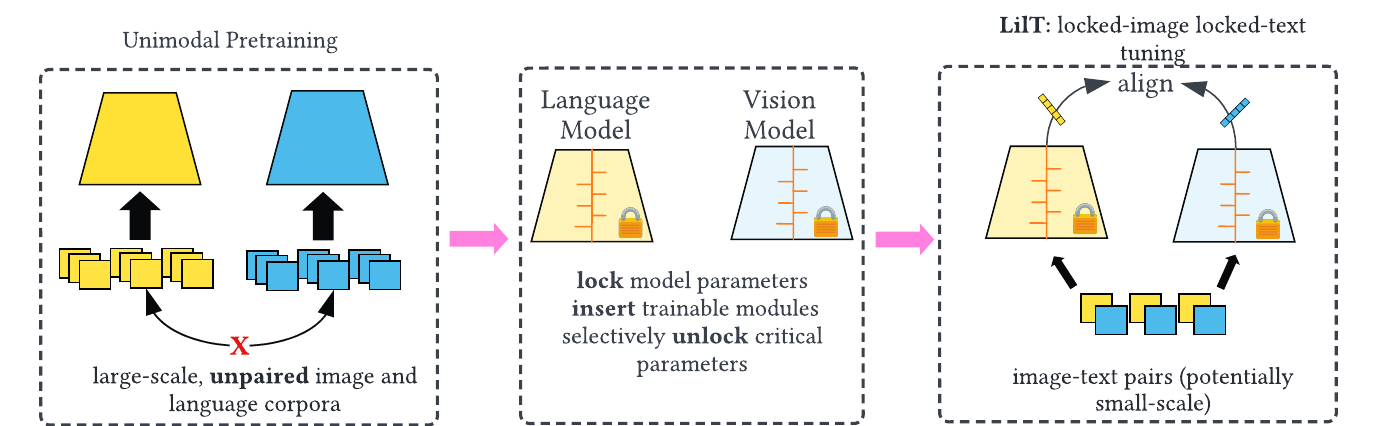}
  \caption{A conceptual diagram. After unimodal pretraining, parameter-efficient transfer to contrastive vision-language alignment is achieved by changing as few as $0.3$\% of the parameters from initialization, matching the performance of full model training.}
  \label{fig:conceptual}
\end{figure}

\section{Methods}
\label{sec:methods}
The basic idea of our approach is to align a vision model and a language model by updating a small percentage of their parameters by gradient descent. 
This involves four main elements.
First, the vision and language model must initialized from strong, pretrained vision and language models, rather than random initialization.
Second, we lock all the parameters in each model.
Third, we selectively unlock critical parameters.
Fourth, we insert small trainable modules into each model to aid adaptation.
There are multiple ways of implementing these strategies, which we cover in this section.
\subsection{Background}
\label{sec:background}
In this section, we briefly cover the mechanics of contrastive language image alignment as used by \citep{clip}, as well as the common "two-tower" \citep{lit}, dual transformer encoder architectures employed by CLIP-style models.
Contrastive language image alignment pulls representations of matched image-text pairs together, while pushing those of unmatched pairs apart. 
The goal is to learn an image encoder $f_{\theta}$ and a text encoder $g_\phi$ such that given an image-text pair $(\mathbf{x}^I, \mathbf{x}^T)$, the encoded representations $f_{\theta}\left(\boldsymbol{x}^{I}\right)$ and $g_{\phi}\left(\boldsymbol{x}^{T}\right)$ are close under a distance metric if they are semantically similar and far apart if not.
Let $\left\{x_{k}^{I}, x_{k}^{T}\right\}_{k=1}^{b}$ be a batch of $b$ image-text pairs. 
For each image $\boldsymbol{x}_{k}^{I}$ in an image-text pair $\left\{x_{k}^{I}, x_{k}^{T}\right\}$, the matched text $\boldsymbol{x}_{k}^{T}$ is the positive, while all other texts within the batch are used as negatives.
The image-to-text contrastive loss $\mathcal{L}_{k}^{I}$ for $\boldsymbol{x}_{k}^{I}$ is then 
$$
\mathcal{L}_{k}^{I}\left(\boldsymbol{x}_{k}^{I},\left\{\boldsymbol{x}_{j}^{T}\right\}_{j=1}^{b}\right)=-\frac{1}{b} \log \frac{\exp \left(s_{k, k}^{I}\right)}{\sum_{j} \exp \left(s_{k, j}^{I}\right)},
$$
where $s_{k, j}^{I}$ is the similarity of the $k$-th image to the $j$-th text.
The similarity function is usually taken to be the cosine similarity, which can be easily computed as $f_{\theta}\left(\boldsymbol{x}^{I}\right)\cdot g_{\phi}\left(\boldsymbol{x}^{T}\right)$ if the representations are normalized to unit length.
Conversely, the text-to-image contrastive loss for $\boldsymbol{x}_{k}^{T}$ is 
$$
\mathcal{L}_{k}^{T}\left(\boldsymbol{x}_{k}^{T},\left\{\boldsymbol{x}_{j}^{I}\right\}_{j=1}^{b}\right)=-\frac{1}{b} \log \frac{\exp \left(s_{k, k}^{T}\right)}{\sum_{j} \exp \left(s_{j, k}^{T}\right)}.
$$
The complete training loss then becomes
\begin{equation}
\mathcal{L}=\frac{1}{2} \sum_{k=1}^{b}\left(\mathcal{L}_{k}^{I}+\mathcal{L}_{k}^{T}\right).
\end{equation}

Architectures for contrastive language image alignment must encode both texts and images to vector representations.
This is usually implemented using separate text encoder and image encoders. 
A variety of choices are possible for these encoders, but we restrict ourselves to the popular \citep{clip,albef,declip,filip,simla,lit,triple_contrastive_learning,ufo} choice of transformer \citep{attention_is_all_you_need} architectures, specifically, the BERT \citep{bert} family of language models for the text encoder, and the ViT \citep{vit} family for the image encoder. 
Let $t(\cdot)$ denote an arbitrary architecture from one of the above families.
After consuming an input $\mathbf{x}$, the transformer $t(\cdot)$ produces a sequence of vectors $t(\mathbf{x})=\left\{\mathbf{z}_{\mathrm{cls}}, \mathbf{z}_{1}, \ldots, \mathbf{z}_{N}\right\}$, where $\mathbf{z}_{\text {cls }}$ is the embedding of the \texttt{[CLS]} token, which is taken to be the representation of the input $\mathbf{x}$ following dimensionality reduction by a trainable linear projection. 
\subsection{Adding Adapters}
\label{sec:growing-adapter}
Aligning the representations of a language transformer and a vision transformer is typically done by updating $100$\% of the parameters in one \citep{lit} or both \citep{clip} of the transformers.
By freezing the transformers, we exclude full-model training, and must use an alternative strategy to align the image and text representations.
A promising approach is inserting a small (relative to each transformer), trainable module into the frozen, pretrained transformers that can learn to modify the internal representations of the transformer it is placed within, such that the representation spaces of the frozen vision and language transformers become aligned while leaving the pretrained parameters untouched. 
We explore two such modules: layerwise adapters \citep{Houlsby2019ParameterEfficientTL,He2021TowardsAU} and "deep" adapters.

\begin{figure}
  \centering
  \includegraphics{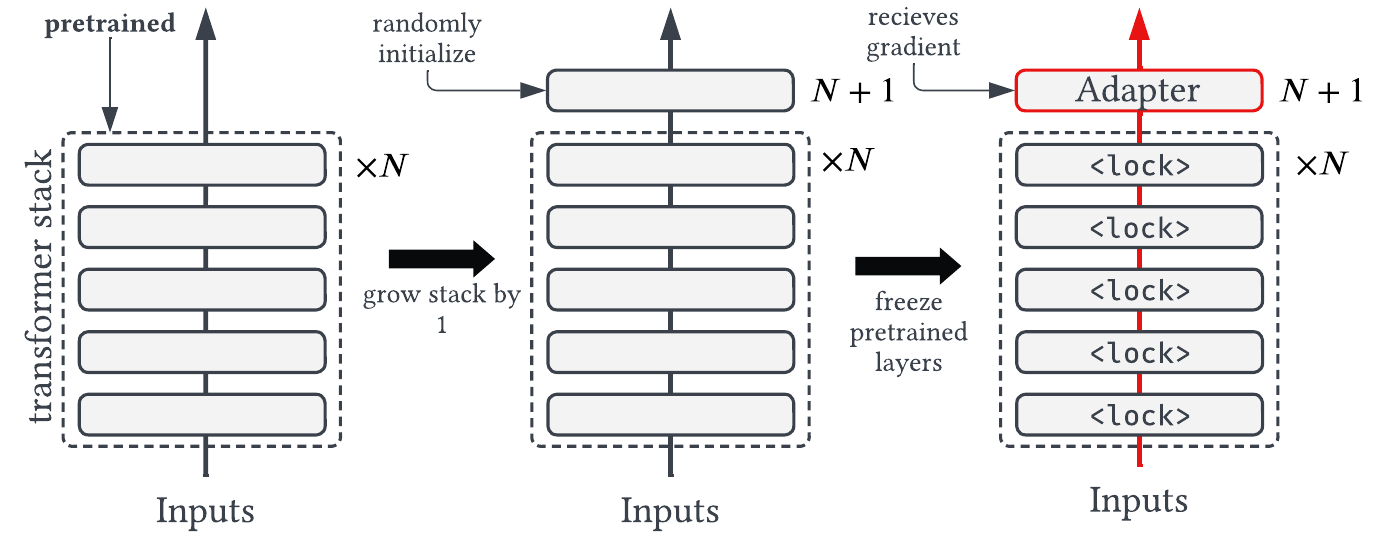}
  \caption{Growing the transformer encoder stack to add a trainable deep adapter to a locked model. The deep adapter is architecturally identical to a layer from the encoder stack.}
  \label{fig:growing-adapter}
\end{figure}

\begin{wrapfigure}{R}{0.5\textwidth}
	\begin{minipage}{0.5\textwidth}
			\vspace{-1em}
			\centering
			\includegraphics[scale=0.8]{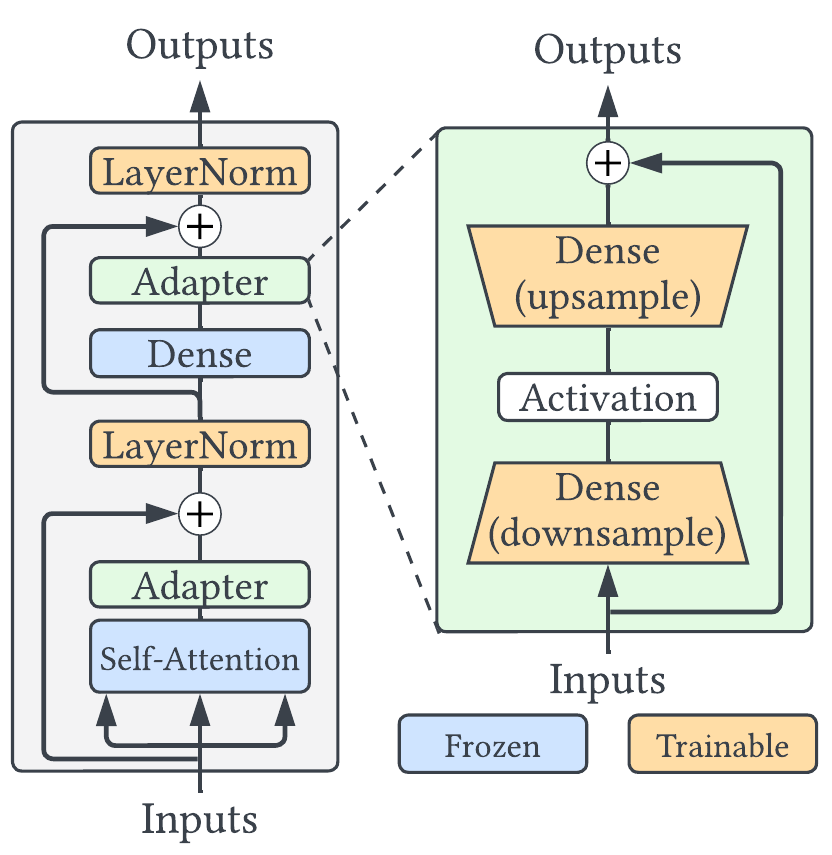}
            \vspace{-3mm}
			\caption{The architecture and placement of layerwise adapters combined with a layernorm unlocking strategy. }
			\label{fig:layerwise-adapter}
	\end{minipage}
\end{wrapfigure}

Layerwise adapters \citep{Houlsby2019ParameterEfficientTL} have been used to adapt pretrained transformer-based language models to new tasks while only updating $2-3$\% of model parameters.
A layerwise adapter is inserted before each layer normalization \citep{Ba2016LayerN} layer in a transformer, and consists of a weight matrix that downsamples the input, followed by an activation function (we use GELU \citep{gelu}) and a weight matrix that restores the input to the original dimensionality, and finally, a residual connection.
We depict the architecture / placement of layerwise adapters in Fig \ref{fig:layerwise-adapter}.

Another solution is to treat the frozen encoders as feature extractors, and learn trainable adapters that align the frozen image and text features.
Transformer architectures can be seen as a stack of identically structured transformer encoder layers, so a natural solution to the problem of designing a trainable adapter atop a stack of frozen transformer encoder layers is to grow the stack, and keep the newly added layers trainable. 
This yields a generic approach (Fig. \ref{fig:growing-adapter}) to add a trainable adapter to a frozen transformer from any of the standardized families (e.g. BERT \citep{bert}, ViT \citep{vit}) that only requires a small number of parameters to recieve gradients ($\approx7$\% for \texttt{bert-base}).
\subsection{Unlocking Parameters}
\label{sec:unlocked-layernorms}
We try two strategies for selectively unlocking parameters in a frozen transformer: unlocking the layer normalization \citep{Ba2016LayerN} parameters, and BitFit \citep{bitfit}. 
Standard transformers \citep{attention_is_all_you_need} have two layer normalization \citep{Ba2016LayerN} modules for each transformer encoder layer, and these are known to play an important role (\S\ref{sec:related}).
Each layer normalization layer has learnable scale $\gamma$ and bias parameters $\beta$ that apply an elementwise scale and shift to the input of the layer normalization layer.
In the first strategy, we allow the layer normalization layers to remain unlocked and receive gradient updates. 
In BitFit \citep{bitfit}, (Bias-term Finetuning), the additive bias terms of \textit{every} module in a transformer encoder layer are allowed to remain unlocked and receive gradient updates.
Both of these strategies unlock a small percentage (0.24\% and 0.31\% of the parameters in a 12-layer \texttt{base} transformer respectively).

\subsection{Implementation Details}
\label{sec:implementation-details}
\textbf{Datasets} We draw $591,753$ image-text pairs from the training set of COCO2014\cite{coco}, following the split of \citet{karpathy_split}.  
The weights of the vision encoders are initialized from DeiT \citet{deit}, and the text encoders are initialized from SimCSE \citep{simcse}.
We train each model with a batch size of 512 on 4x NVIDIA A6000 GPUs for 15 epochs, using the AdamW optimizer \citep{adamw} optimizer with a weight decay of 0.02. 
The learning rate is warned up to 1$e^{-4}$ in the first 10 epochs, and then decayed to $1e^{-5}$.
We use random crops of resolution $256 \times 256$ with RandAugment\citep{randaugment}, with colors transformations removed following \cite{albef}.

\section{Experiments}
\label{sec:experiments}
We conduct experiments on zero-shot multimodal classification, image-text retrieval, and multilingual image text retrieval to investigate the following research questions.
\begin{enumerate}
    \item Can contrastive vision language models be created through parameter-efficient transfer learning?
    \item How do different methods for parameter efficient transfer learning interact with each other?
    \item Do contrastive vision language models created through parameter-efficient transfer learning conserve useful knowledge from their initializations better than full-model finetuning?
    \item Does parameter-efficient transfer learning scale with respect to model size and dataset size?
\end{enumerate}

We evaluate all models on five tasks: zero-shot natural-language guided image classification \citep{clip}, image-to-text retrieval (TR), text-to-image retrieval (IR), and 0-shot TR/IR.
For zero-shot classification, we use the ImageNetV2 \citep{imagenetv2} test set.
For IR/TR, we use the COCO2014 test split of \citet{karpathy_split}, containing 5k images and 25k captions.
For zero-shot IR/TR, we use the test set of Flickr30k\citep{flickr30k}, containing 1k images and 5k captions.


\subsection{Ablation Study}
\label{sec:ablations}
\begin{table}[]
    \centering
        \caption{An ablation study with \texttt{bert-base} as the text encoder and a ViT-B/16 as the image encoder. An \emoji{ice} indicates the component is locked and does not recieve gradient updates, while \emoji{fire} indicates the opposite.
        LN(\emoji{ice} $_T$/\emoji{fire} $_I$) indicates the layer normalization weights in the text encoder were locked while those of the image encoder recieved gradient updates, and vice versa for LN(\emoji{fire}$_T$/\emoji{ice}$_I$). $\theta$ is the trainable linear projection. TR and IR is mean text retrieval and image retrieval scores across Rank-1,5,10. Deep (Fig \ref{fig:layerwise-adapter}) and Layerwise (Fig. \ref{fig:growing-adapter}) adapters are detailed in \S\ref{sec:growing-adapter}, and BitFit in \S\ref{sec:unlocked-layernorms}.
        }
    \label{tab:ablations}
    \begin{adjustbox}{max width=\textwidth}
        \begin{tabular}{llcccccllll}
\toprule
& & \multicolumn{5}{c}{Components} &  & \multicolumn{2}{c}{Flickr}  & \multicolumn{1}{c}{ImageNet V2} \\
\cmidrule(l{0.5em}r{0.5em}){3-7}  \cmidrule(l{0.5em}r{0.5em}){9-11}
 & & TE & IE & $\theta$ & Unlock Strategy & Adapter & \% Trained  & TR & IR & Acc-1 \\
\midrule 
(a) & Frozen & \emoji{ice} & \emoji{ice} & \emoji{ice} & LN(\emoji{ice}$_T$/\emoji{ice}$_I$) & - & 0.00 \% & 0.8 & 1.3 & 0.2 \\
(b) & LN Only & \emoji{ice} & \emoji{ice} & \emoji{ice} & LN(\emoji{fire}$_T$/\emoji{fire}$_I$) & - & 0.04 \% & 24.3 & 21.6 & 4.3 \\
(c) & Projection Only & \emoji{ice} & \emoji{ice} & \emoji{fire} & LN(\emoji{ice}$_T$/\emoji{ice}$_I$) & - & 0.20\% & 38.7 & 31.8 & 6.7 \\
(d) & LilT$_{LN}$ & \emoji{ice} & \emoji{ice} & \emoji{fire} & LN(\emoji{fire}$_T$/\emoji{fire}$_I$) & - & 0.24\% & 62.3 & 51.74 & 12.5 \\
(e) & LilT$_{BF}$ & \emoji{ice} & \emoji{ice} & \emoji{fire} & BitFit & - & 0.31\% & 62.6 & 52.1 & 12.6 \\
(f) & LilT$_{DA}$ w/o LN & \emoji{ice} & \emoji{ice} & \emoji{fire} & LN(\emoji{ice}$_T$/\emoji{ice}$_I$) & Deep & 6.96 \% & 57.5 & 47.8 &  9.02 \\
(g) & LilT$_{DA}$ & \emoji{ice} & \emoji{ice} & \emoji{fire} & LN(\emoji{fire}$_T$/\emoji{fire}$_I$) & Deep & 6.99 \% & 68.6 & 58.5 &   12.9 \\
(h) & LilT$_{LwA}$ w/o LN & \emoji{ice} & \emoji{ice} & \emoji{fire} & LN(\emoji{ice}$_T$/\emoji{ice}$_I$) & Layerwise & 6.97 \%  & 74.8 & 63.9 &   12.0 \\
(i) & LilT$_{LwA}$ & \emoji{ice} & \emoji{ice} & \emoji{fire} & LN(\emoji{fire}$_T$/\emoji{fire}$_I$) & Layerwise & 7.01 \%  & 75.4 & 64.4 &   12.2 \\
(j) & LilT$_{LwA}$(BitFit)& \emoji{ice} & \emoji{ice} & \emoji{fire} & BitFit & Layerwise & 7.09\% & 75.3 & 64.4 &   12.2 \\
(k) & LilT$_{DA}$ (BitFit) & \emoji{ice} & \emoji{ice} & \emoji{fire} & BitFit & Deep & 7.06\%  & 68.7 & 58.4 &  13.2 \\
\midrule
(l) & LiT &\emoji{fire} & \emoji{ice} & \emoji{fire} & LN(\emoji{fire}$_T$/\emoji{ice}$_I$) & - & 56.01 \% & 66.1 & 53.5 & 15.0 \\
(m) & LiT (reversed) & \emoji{ice} & \emoji{fire} & \emoji{fire}& LN(\emoji{ice}$_T$/\emoji{fire}$_I$) & - & 43.99 \% & 53.7 & 46.22 & 8.8 \\
(n) & LiT + LilT$_{DA}$ & \emoji{fire} & \emoji{ice} & \emoji{fire}& LN(\emoji{fire}$_T$/\emoji{fire}$_I$) & Deep & 65.87 \% & 84.2 & 75.2 & 13.6 \\
(o) & LiT + LilT$_{LwA}$ & \emoji{fire} & \emoji{ice} & \emoji{fire}& LN(\emoji{fire}$_T$/\emoji{fire}$_I$) & Layerwise & 57.57\% & 76.7 & 64.9 & 13.84 \\
\midrule
(p) & CLIP & \emoji{fire} & \emoji{fire} & \emoji{fire}& LN(\emoji{fire}$_T$/\emoji{fire}$_I$) & - & 100.0 \% & 75.8 & 65.8 & 12.3 \\
\bottomrule
\end{tabular}
    \end{adjustbox}
\end{table}
The results of the study are displayed in Table \ref{tab:ablations}.
After updating only $0.24$\% of parameters, parameter unlocking methods achieve equivalent zero-shot classification performance to full-model training: compare (d) \& (e) to (p).
However, parameter unlocking alone is insufficient to achieve the image-text retrieval abilities of full-model training, but adapter-based methods (f-k) can match full-model training (p) in both zero-shot classification and image-text retrieval. 
BitFit and layer normalization unlocking are interchangeable as parameter unlocking strategies ($<0.2$\% difference between (f/j) and (h/i)).
LilT$_{LwA}$ (h), with the layerwise adapters, is substantially better ($\approx7\%$) at image text retrieval than LilT$_{DA}$ (f), and only slightly worse at classification.
LilT and LiT are complimentary (m/n), and it is possible to only align only \textit{one} of the encoders in a parameter-efficient manner.
While LiT (k) excels at image classification, it suffers from a similar problem as parameter unlocking strategies: it is relatively poor at image text retrieval.

\textbf{Discussion} First, it is clear that creating contrastive vision-language models through parameter-efficient transfer learning is feasible, and there are clear differences between model capabilities induced by different parameter-efficient transfer learning methods.
Layerwise adapters stand out as the parameter-efficient transfer learning strategy capable of matching or exceeding full-model training.
However, in cases where the language distribution is sufficiently simple (e.g. a list of singular words), parameter unlocking is sufficient, and easier to implement.
Deep adapters stand out for their ability to achieve \textit{better} performance than full-model training when combined with LiT (m).

\subsection{Conservation of knowledge from initialization}
We hypothesize that parameter efficient transfer learning preserves more knowledge from initialization than full model finetuning, and this is beneficial in some realistic scenarios. 
Low-resource languages likely do not have large-scale image-text pairs available to train a multimodal CLIP-like model for that language. 
However, \textit{unimodal}, multilingual language models that have been trained on a dataset containing sentences from a given low-resource language often exist. 
A possible solution in this situation is to train a CLIP-like model on available image-text pairs from a high-resource language, while using a multilingual language model as the text encoder.
The resulting model may be able to generalize to image-text retrieval tasks in a language unseen during vision-language alignment due to the multilinguality of the pretrained text encoder.
We simulate this setting by aligning a pretrained multilingual \texttt{BERT-base} model with an ImageNet-pretrained ViT-B/16 on English-only image-text pairs, and evaluate it on image-text pairs in six different languages that the model was never provided paired images for. 
If parameter-efficient training preserves more knowledge from initialization, and that knowledge is useful, we expect that the retrieval model created through parameter efficient transfer learning should retain more of its multilingual language ability, and hence display greater accuracy on non-English languages.

\begin{table}[]
    \centering
        \caption{Cross-lingual zero-shot retrieval. A multilingual \texttt{bert-base} model is aligned with a ViT-B/16 on English image-text pairs from COCO, and evaluated on image-text pairs in languages unseen during alignment.}
    \label{tab:xlingual}
    \begin{adjustbox}{max width=\textwidth}
\begin{tabular}{lllllllllllllll}
\toprule
{} & \multicolumn{2}{c}{RU} & \multicolumn{2}{c}{PL} & \multicolumn{2}{c}{TR} & \multicolumn{2}{c}{ZH} & \multicolumn{2}{c}{KO} & \multicolumn{2}{c}{IT} & \multicolumn{2}{c}{ES} \\
\cmidrule(l{0.5em}r{0.5em}){2-3} \cmidrule(l{0.5em}r{0.5em}){4-5} \cmidrule(l{0.5em}r{0.5em}){6-7} \cmidrule(l{0.5em}r{0.5em}){8-9} \cmidrule(l{0.5em}r{0.5em}){10-11} \cmidrule(l{0.5em}r{0.5em}){12-13} \cmidrule(l{0.5em}r{0.5em}){14-15}
{} &     TR &     IR &     TR &     IR &     TR &     IR &     TR &     IR &     TR &     IR &     TR &     IR &     TR &     IR \\
\midrule
LiT     &  45.17 &  40.17 &   44.0 &  41.83 &  24.17 &  23.33 &  64.67 &   61.0 &  34.17 &  29.67 &  60.17 &   56.0 &  65.67 &  62.33 \\
CLIP     &  57.67 &  53.17 &  59.17 &  54.83 &  33.33 &  29.83 &   79.0 &   \textbf{74.0} &  42.33 &  35.33 &   71.0 &  65.33 &  75.67 &   69.5 \\
\midrule
LilT$_{DA}$  &   58.5 &  51.33 &  60.33 &  55.33 &  42.33 &   35.0 &  74.17 &  67.67 &  44.67 &  35.67 &   74.5 &  68.83 &   77.0 &  74.17 \\
LilT$_{LwA}$ &  \textbf{61.83} &   \textbf{57.0} &   \textbf{63.0} &   \textbf{56.5} &   \textbf{46.5} &   \textbf{41.0} &   \textbf{79.0 }&  72.83 &   \textbf{50.0} &  \textbf{43.67} &  \textbf{77.67} &  \textbf{72.17} &  \textbf{79.17} &   \textbf{74.5} \\
\midrule
$\Delta$    &   $\uparrow$4.17 &   $\uparrow$3.83 &   $\uparrow$3.83 &   $\uparrow$1.67 &  $\uparrow$13.17 &  $\uparrow$11.17 &    $\uparrow$0.0 &  $\uparrow$-1.17 &   $\uparrow$7.67 &   $\uparrow$8.33 &   $\uparrow$6.67 &   $\uparrow$6.83 &    $\uparrow$3.5 &    $\uparrow$5.0 \\
\bottomrule
\end{tabular}
\end{adjustbox}
\end{table}
We reuse the English training data from \S\ref{sec:implementation-details}, and evaluate each model on the test set of \citet{aggarwal2020zeroshot}, which contains 1400 image-text pairs, split equally between Russian, Polish, Turkish, Chinese, Korean, Italian, and Spanish.
We summarize results in Table \ref{tab:xlingual}.
LilT$_{LwA}$ outperforms CLIP on $12/14$ tasks (5.3\% absolute improvement), while LilT$_{DA}$ achieves better performance than CLIP on 11/14 tasks (1.4\% absolute improvement).
This suggests that parameter-efficient transfer learning conserves more information from initialization, and that information is  useful for multimodal tasks.


\subsection{Scaling with respect to data and model size}
Can parameter-efficient transfer learning take advantage of larger models and larger amounts of data? 
We test the the performance of parameter-efficient transfer learning as the amount of image-text pairs is increased to 1500k from 591k (Table \ref{tab:scaling-1500k}) and as model size is increased (Table \ref{tab:performance}) from \texttt{base} ($\approx 200$M params) to \texttt{large} ($\approx 700$M params).
When the amount of training pairs available triples, parameter-efficient transfer learning continues to match the performance of full-model training: (b) vs (d) in Table \ref{tab:scaling-1500k}.
Similarly, the performance of parameter-efficient transfer learning improves as model size increases: (a) vs (b) \& (c) vs (d) in Table \ref{tab:performance}.
\begin{table}[]
    \centering
        \caption{Zero-shot task performance of \texttt{base/large} models after parameter-efficient training.$LwA$/$DA$ indicates adapter types, corresponding to (rows h/f in Table \ref{tab:ablations}). }
    \label{tab:performance}
    \begin{adjustbox}{max width=\textwidth}
        \begin{tabular}{lccccccccc}
\toprule
\multicolumn{4}{c}{Model (591k Training Pairs)} &  & \multicolumn{2}{c}{Flickr} & & \multicolumn{2}{c}{ImageNet V2} \\
\cmidrule(l{0.5em}r{0.5em}){2-4}  \cmidrule(l{0.5em}r{0.5em}){5-8} \cmidrule(l{0.5em}r{0.5em}){9-10}
& Configuration & \# Trainable & \% Trained &  TR@1 & IR@1 & TR@5 & IR@5 & Acc-1 & Acc-5 \\
\midrule
 (a) &   LilT$_{DA}$-base & 14.65 M & 7.51\% & 47.6 & 34.46 & 74.1 & 64.92 & 12.94 & 28.39  \\
 (b) &   LilT$_{DA}$-large & 25.92 M & 4.06\% & 57.6 & 42.18 & 82.2 & 72.38 & 13.97 & 30.89  \\ \midrule
  (c) &   LilT$_{LwA}$-base & 14.67 M & 7.01\% & 56.8 & 41.7 & 81.1 & 70.74 & 12.18 & 27.78  \\
   (d) &   LilT$_{LwA}$-large & 51.18 M & 7.43\% & 63.5 & 50.7 & 88.5 & 79.14 & 14.05 & 31.31 \\
 \midrule
 (e) & LiT-base & 109.28 M & 56.01\% & 44.1 & 29.64 & 72.1 & 59.94 & 15.0 & 29.44 \\
 (f) & CLIP-base & 195.13 M & 100.0\% & 56.1 & 44.3 & 81.7 & 71.98 & 12.29 & 28.44  \\
\bottomrule
\end{tabular}
    \end{adjustbox}
\end{table}
\begin{table}[]
    \centering
        \caption{Zero-shot performance of \texttt{base} models after larger-scale pretraining (1.5M pairs).}
    \label{tab:scaling-1500k}
    \begin{adjustbox}{max width=\textwidth}
        \begin{tabular}{lccccccccc}

\toprule

\multicolumn{4}{c}{Model (1.5M Pairs)} &  & \multicolumn{2}{c}{Flickr} & & \multicolumn{2}{c}{ImageNet V2} \\
\cmidrule(l{0.5em}r{0.5em}){2-4}  \cmidrule(l{0.5em}r{0.5em}){5-8} \cmidrule(l{0.5em}r{0.5em}){9-10}
& Configuration & \# Trainable & \% Trained &  TR@1 & IR@1 & TR@5 & IR@5 & Acc-1 & Acc-5 \\
\midrule

 (a) & LiT-base & 109.28 M & 56.01\% & 48.8 & 32.72 & 78.1 & 63.02 & \textbf{20.63} & \textbf{38.12} \\
 (b) & CLIP-base & 195.13 M & 100.0\% & 60.5 & 43.8 & 84.7 & 72.16 & 16.61 & 35.14  \\
 \midrule
  (c) &   LilT$_{DA}$-base & 14.65 M & 7.51\% & 50.4 & 35.66 & 78.2 & 65.3 & 16.98 & 35.53  \\
  (d) &   LilT$_{LwA}$-base & 14.67 M & 7.01\% & \textbf{61.1} & \textbf{44.5} & \textbf{85.6} & \textbf{72.9} & 15.83 & 35.31  \\

\bottomrule
\end{tabular}
    \end{adjustbox}
\end{table}

\subsection{What happens during alignment?}
\begin{figure}
  \centering
  \includegraphics{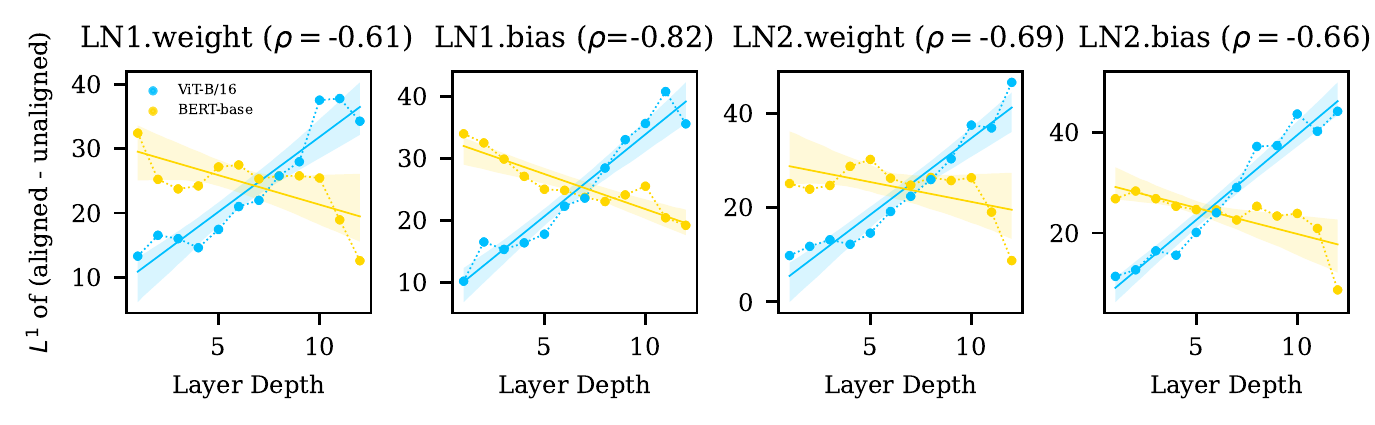}
  \caption{The depth of the layer normalization layers affects how much they are changed by alignment training, and the pattern is reversed between the image and text encoders. $\rho$ is the Pearson correlation coefficient, and the translucent blue/yellow shading indicates 95\% confidence intervals. }
  \label{fig:layernorm-analysis}
\end{figure}
\begin{figure}
  \centering
  \includegraphics{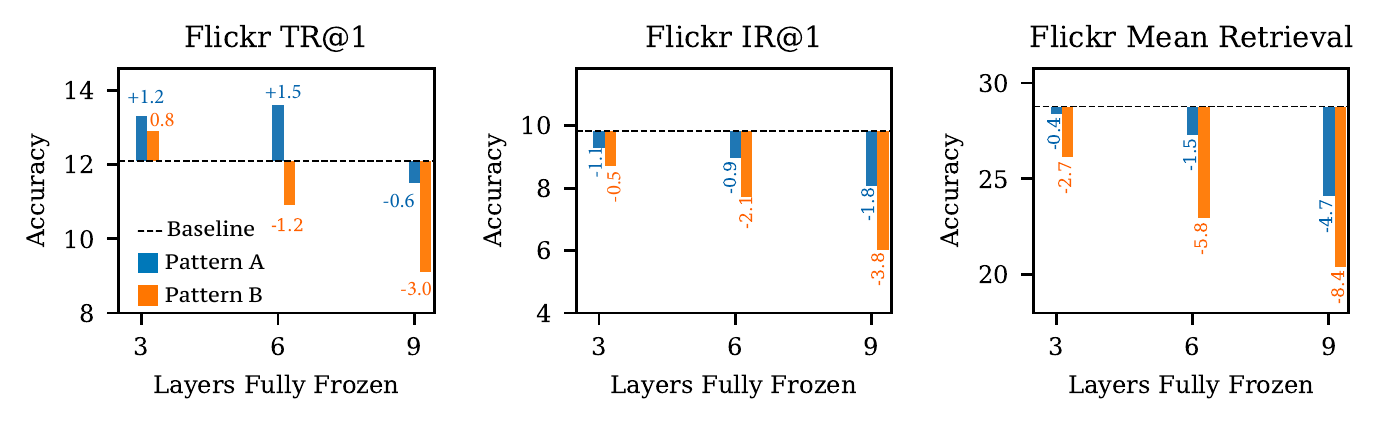}
  \caption{We freeze all parameters except for the LN parameters, then progressively lock LN parameters by layer. Fig \ref{fig:layernorm-analysis} suggests that freezing the LN parameters in the deepest layers of the language model and the shallowest layers of the vision model (Pattern A) should have a smaller effect on performance than the opposite pattern (Pattern B), relative to the baseline (LNs in every layer unlocked) which we observe.}
  \label{fig:layernorm-consequences}
\end{figure}

We attempt to understand how alignment changes the language and vision model by studying the layer normalization layers of each model.
Let $f_{\theta}$ be an image encoder $g_\phi$ be a text encoder.
We initialize $f_{\theta}$ with weights from DEiT\cite{deit}, and $g_\phi$ with weights from SimCSE \cite{simcse}.
We then lock all parameters except the layer normalization layers (configuration (c) in Tab. \ref{tab:ablations}), and train the model following the standard CLIP training procedure, resulting in a pair of aligned encoders ($\bar{f}_{\theta}$, $\bar{g}_\phi$).
In total, we have four different models: the unaligned and aligned image encoders ($f_{\theta},\bar{f}_{\theta}$) and the unaligned and aligned text encoders ($g_\phi, \bar{g}_\phi$).
Without loss of generality, we describe our procedure for the text encoder pair ($g_\phi, \bar{g}_\phi$).
Let $\mathrm{LN}^1_i(\gamma, \beta)$ and $\mathrm{LN}^2_i(\gamma, \beta)$, denote the two normalization sublayers of the $i$-th layer in the transformer encoder stack.
For layer $i \in {1,2, \ldots N}$, we plot the $L^1$ norm of the difference between the trainable layer normalization parameters $\gamma, \beta$ of the aligned and unaligned encoders. 
We plot the results in Fig \ref{fig:layernorm-analysis}.
Surprisingly, the text and image encoders display clearly opposite patterns (negative Pearson's \textit{r}). 
In the text encoder, the difference between the aligned and unaligned layer normalization parameters decreases with depth --- layer normalization parameters in the deeper layers of the text encoder change less as a result of alignment training. 
This is the \textit{opposite} of the image encoder. 
In the image encoder, the layer normalization parameters which shift the most as a result of training are the deepest.
We conduct another experiment with 50k pairs (Fig \ref{fig:layernorm-consequences}) to test the consequences of this pattern.

\textbf{Discussion} The patterns in the layer normalization layers may indicate that during alignment, the language and image modalities undergo changes at different semantic levels. The shallowest three layer normalization layers of the ViT-B/16 experience a $\approx 70\%$ lower magnitude shift than the deepest three layers. The shallow layers of a vision transformer attend more to local information \citep{Raghu2021DoVT}, while the deeper layers attend more to global context. Intuitively, this makes sense -- we should expect an asymmetry between the amount of information in a short image caption compared to a dense image. Simple natural language concepts are often visually complex. Interestingly, this has already been exploited by certain vision-language models --- \citep{simla,albef}  align \textit{the lower half} of their text encoder to the visual encoder, while using the top half for a different purpose. This makes sense, given that the lower layers of the text encoder seem to change the most during alignment.

\section{Related Work}
\label{sec:related}
\textbf{Vision-Language Pretraining} The dual-encoder CLIP \citep{clip} (400m pairs) and ALIGN \citep{align} (1b+ pairs) architectures were the first attempts at large-scale contrastive image-language alignment using the InfoNCE \citep{infonce} loss to maximize the mutual information between matched image and text pairs.
Subsequent work \citep{basic,declip,filip,defilip,triple_contrastive_learning,simla,albef} has improved on the training tasks, dataset, and architecture of CLIP.
While systems utilizing a multimodal encoder and cross attention \cite{blip,simla,Wang2022UnifyingAT,Lu2022UnifiedIOAU,Zhu2021UniPerceiverPU} perform better on benchmarks, their multimodal encoder makes them unsuitable for latency-sensitive search application, because rather than learning separate but aligned image and text embeddings, they learn a single multimodal embedding for an image-text pair.
Thus, neural search remains the domain of contrastive vision-language models.

\textbf{Frozen Language Models}
\citet{NEURIPS2021_01b7575c} demonstrated that pretrained large language models are capable of quickly adapting to image understanding.
They use an autoregressive transformer-based language model, which is frozen.
A trainable ResNet \citep{resnet} is then trained to transform images into input the frozen transformer can understand, by backpropagating the loss through the frozen transformer.
MAGMA \cite{Eichenberg2021MAGMAM}, FROMAGE \cite{GroundingLanguageModelsFried2023} and FLAMINGO \cite{Alayrac2022FlamingoAV} scaled the conceptual approach of \citet{NEURIPS2021_01b7575c} to billions of parameters, and recently, \citet{LinearlyMappingImagePavlick2022} have shown that a simple linear mapping is enough to allow a frozen large language model to (roughly) understand visual input, as long as the visual encoder has been trained to represent visual concepts aligned to language (e.g. CLIP).
However, emerging approaches such as BLIP-2 \cite{Li2023BLIP2BL} show that by combining soft prompting with a frozen LLM and a trainable visual encoder, a LLM can achieve state-of-the-art accuracy on visuolinguistic understanding tasks such as visual question answering.
\citet{lu2021fpt} propose the idea that transformers trained on language are capable of a form of universal computation, and can adapt to new tasks even if they are frozen, and do so better than fine-tuned models. 
However, \citet{Rothermel2021DontSY} find the findings may be reversed under certain hyperparameter settings.
Interestingly, both note that the normalization layers seem to play an important role in this adaptation. 

\textbf{Parameter-Efficient Finetuning}
Many forms of adapters \citep{adapters,hyperformer,compacter} have been explored in natural language processing.
VL-Adapter \citep{Sung2021VLAdapterPT} investigate adapters in vision-language, but assume aligned visual representations.
\citet{Lester2021ThePO} find that for very large language models, parameter-efficient adaptation approaches such as soft prompting are equivalent to finetuning the large language model. 
\citet{Liu2021PTuningVP} extend this finding, showing that combining soft prompting with adapters can often \textit{exceed} finetuning on a given downstream task.
Both prefix \citep{Li2021PrefixTuningOC} and prompt \citep{Lester2021ThePO} tuning can also be understood as exploiting the knowledge in frozen transformers, as their optimization loops involve freezing the language model, effectively turning it into a part of the loss.
\citet{DBLP:conf/nips/ZhangH20} develop a training scheme that progressively unfreezes / freezes layers of a transformer language model, and see significant improvements in training speed. 
Progressive growth approaches \citep{Gu2021OnTT} slowly increase the depth of a transformer as training proceeds.

\textbf{Layer Normalization in Transformers} \citet{Kovaleva2021BERTBO} find that the representations of transformers contain outlier dimensions that disrupt the quality of the learned embedding, and point to high-magnitude parameters in the layer normalization layers.
A variety of techniques targeting layer normalization in transformers have been proposed, with various benefits.
\citet{Xiong2020OnLN} prove that the placement of layer normalization layers relative to the residual connection in the transformer block contributes to learning instability under large learning rates, and propose an alternate placement. 
In contrast, FixUp \citep{Huang2020ImprovingTO} develops a novel initialization scheme for transformers that enables removing the normalization layers entirely. 
ReZero \citep{Bachlechner2021ReZeroIA} adds a learnable gate parameter to each residual connection before layer normalization, and demonstrate training extremely deep transformers quickly. 
\section{Conclusion \& Future Work}
We show that the performance of full model training for contrastive vision language alignment can be matched by updating a small number of parameters in existing vision models and language models, followed by an insertion of trainable modules.
This suggests that the current paradigm of full-model training for contrastive vision language alignment involves significant unnecessary computation, and can be replaced by parameter-efficient transfer learning when the downstream use cases are natural-language classification or image-text retrieval.
Current alignment strategies align representations from the top of each encoder stack. 
We find that in the text encoder, alignment changes the normalization parameters in the shallowest layers the most, while it is the opposite for the image encoder. 
Investigating and exploiting the asymmetry between vision and language could yield further benefits for multimodal understanding or more efficient training strategies.
For future work, it would be interesting to analyze whether CLIP-like models created through parameter-efficient transfer learning are similar to CLIP in ways other than performance --- for example, are they more or less biased? Or more or less robust to distribution shift? 
Another useful line of investigation would be probing vision-language models further to understand how alignment training effects the ability of the model to understand language. 
In summary, we believe that existing training methods are not fully exploiting the knowledge that exists in their initializations.
Our approach presents one simple but effective way to use that knowledge.
\subsubsection*{Acknowledgments}
This work was supported by a faculty award from NEC Laboratories America.

\bibliography{iclr2023_conference}
\bibliographystyle{iclr2023_conference}

\section{Appendix}
\subsection{Additional Datasets}
\FloatBarrier
We conduct zero-shot classification experiments on three further datasets (Table \ref{tab:additional-datasets}): CIFAR-100 \cite{cifar}, SVHN\cite{svhn}, and ImageNet-A\cite{natural_adversarial_examples}.
As CIFAR-100 and SVHN are both standard datasets, we only briefly describe them here. 
The CIFAR-100 dataset consists of 60k 32x32 colour images divided into 100 classes containing 600 images per class. 
Each class has 500 training and 100 test images, for a total of 50k training and 10k test images.
We use the CIFAR-100 test set for the evaluations. 
SVHN is a harder version of MNIST \cite{mnist}, consisting of natural images of digits cropped from street-level pictures.
We use the 26k test images for evaluation.
ImageNet-A consists of \textit{natural adversarial examples} from the ImageNet1k distribution, which are natural, correctly labeled images that classifiers incorrectly classify with high confidence. 
We use the 7k test images. 
\begin{table}[!hbt]
    \centering
        \caption{
        Evaluation on additional zero-shot classification tasks. 
        \textbf{First place} is in bold and \textcolor{red}{second place} is in red. 
        \colorbox{LimeGreen}{LilT} models are boxed in green.
        Acc-1 stands for top-1 accuracy, and Acc-5 is top-5 accuracy.
        Higher is better.
        }
    \label{tab:additional-datasets}
    \begin{adjustbox}{max width=\textwidth}
\begin{tabular}{lccccccccc}
\toprule
\multicolumn{4}{c}{Model} &  \multicolumn{2}{c}{CIFAR100} & \multicolumn{2}{c}{SVHN} & \multicolumn{2}{c}{ImageNet-A} \\
\cmidrule(l{0.5em}r{0.5em}){2-4}  \cmidrule(l{0.5em}r{0.5em}){5-6} \cmidrule(l{0.5em}r{0.5em}){7-8} \cmidrule(l{0.5em}r{0.5em}){9-10}
& Configuration & \# Trainable & \% Trained &  Acc-1 & Acc-5 & Acc-1 & Acc-5 & Acc-1 & Acc-5 \\
\midrule
  \rowcolor{LimeGreen}(a) & LilT-tiny & 736.45 K & 7.37 & 16.98 & 37.49 & \textcolor{red}{13.0} & 57.39 & 2.77 & 9.15  \\
 (b) & LiT-tiny & 4.45 M & 44.57 & 18.33 & 39.14 & 12.47 & 55.02 & 3.39 & 11.03  \\
 \rowcolor{LimeGreen}(c) & LilT-small & 5.19 M & 10.28 & 27.52 & 50.28 & 11.95 & 54.15 & 4.79 & 13.8  \\
 (d) & CLIP-tiny & 9.99 M & 100.0 & 18.74 & 41.1 & \textbf{14.97} & \textbf{63.18} & 2.73 & 10.49  \\
 \rowcolor{LimeGreen}(e) & LilT-base & 14.65 M & 7.51 & \textcolor{red}{29.9} & \textcolor{red}{53.77} & 11.84 & 57.08 & 5.11 & 15.8  \\
 \rowcolor{LimeGreen}(f) & LilT-large & 25.92 M & 4.06 & \textbf{31.33} & \textbf{57.93} & 7.39 & 42.21 & \textbf{7.61} & \textbf{23.44}  \\
 (g) & LiT-small & 28.73 M & 56.98 & 26.88 & 47.17 & 12.3 & \textcolor{red}{59.17} & 5.37 & 16.01  \\
 (h) & CLIP-small & 50.42 M & 100.0 & 26.43 & 49.54 & 7.18 & 54.41 & 4.41 & 14.45  \\
 (i) & LiT-base & 109.28 M & 56.01 & 26.15 & 48.69 & 11.51 & 55.75 & \textcolor{red}{5.92} & \textcolor{red}{18.13}  \\
 (j) & CLIP-base & 195.13 M & 100.0 & 25.25 & 50.93 & 9.47 & 53.33 & 4.68 & 16.41  \\
 \midrule
 (k) & VGG-19\cite{natural_adversarial_examples} & 143M M & 100.0 & - & - & - & - & 2.72 & -  \\
 (l) & ResNet-50\cite{natural_adversarial_examples} & 23 M & 100.0 & - & - & - & - & 2.17 & -  \\
 (m) & ResNet-101 \cite{natural_adversarial_examples} & 44.7 M & 100.0 & - & - & - & - & 4.9 & -  \\
 (n) & ResNet-152\cite{natural_adversarial_examples} & 60.4 M & 100.0 & - & - & - & - & 5.2 & -  \\
\bottomrule
\end{tabular}
\end{adjustbox}
\end{table}
\subsection{Natural Adversarial Examples}
Vision language models display impressive performance on ImageNet-A.
ImageNet-A can be considered a "hard slice" of the ImageNet distribution, containing samples which are problematic for supervised classifiers.
Suprisingly, the zero-shot classification performance of self-supervised vision-language models on ImageNet-A matches and is sometimes greater than the performance of supervised classifiers (ResNet-50 \cite{resnet} and VGG-19 \cite{vgg}).
This may be partially due to the parameter count --- there are more total parameters in most of the vision-language models compared to the supervised CNNs.
However, considering that the vision-language models are facing a harder problem (performing zero-shot classification), their performance relative to supervised CNNs is surprising.
\subsection{Where do the models fail?}
On the SVHN dataset, performance is poor. 
The large models perform worse than random chance ($<10\%$), and the smaller the model, the better it performs. 
One explanation could be that there is no way for the models to learn a correspondence between images of digits and the name of each digit, as nothing similar appears in the COCO training distribution, which only contains common objects.
\subsection{Does pretraining matter?}
\FloatBarrier
\subsubsection{Pretraining vs. Random Initialization}
\begin{figure}[htbp]
    \centering
    \includegraphics{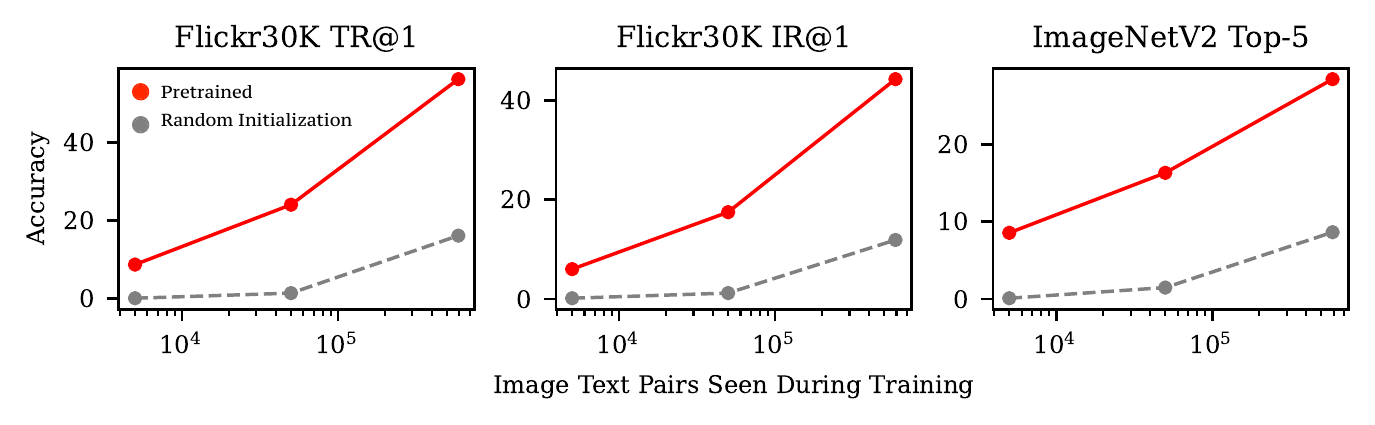}
    \caption{The effect of pretraining on model performance.}
    \label{fig:pretraining-vs-random}
\end{figure}
We follow the standard training procedure  (\S \ref{sec:implementation-details}) and train a CLIP-base model where both of the encoders are initialized randomly, instead of using weights initialized from unimodally pretrained models (DeIT \cite{deit} and SimCSE \cite{simcse}).
We train three models, one for each dataset size.
The results can be seen in Fig \ref{fig:pretraining-vs-random}.
Compared to the randomly initialized model, the pretrained model is substantially better across all three datasets and all 3 model sizes. 
However, it is likely that the benefit of unimodal pretraining will be diminished as the number of training pairs available for multimodal vision-language pretraining increases, although we do not explore this.
\subsubsection{Does the kind of unimodal pretraining matter?}
\FloatBarrier
\begin{figure}[htbp]
    \centering
    \includegraphics{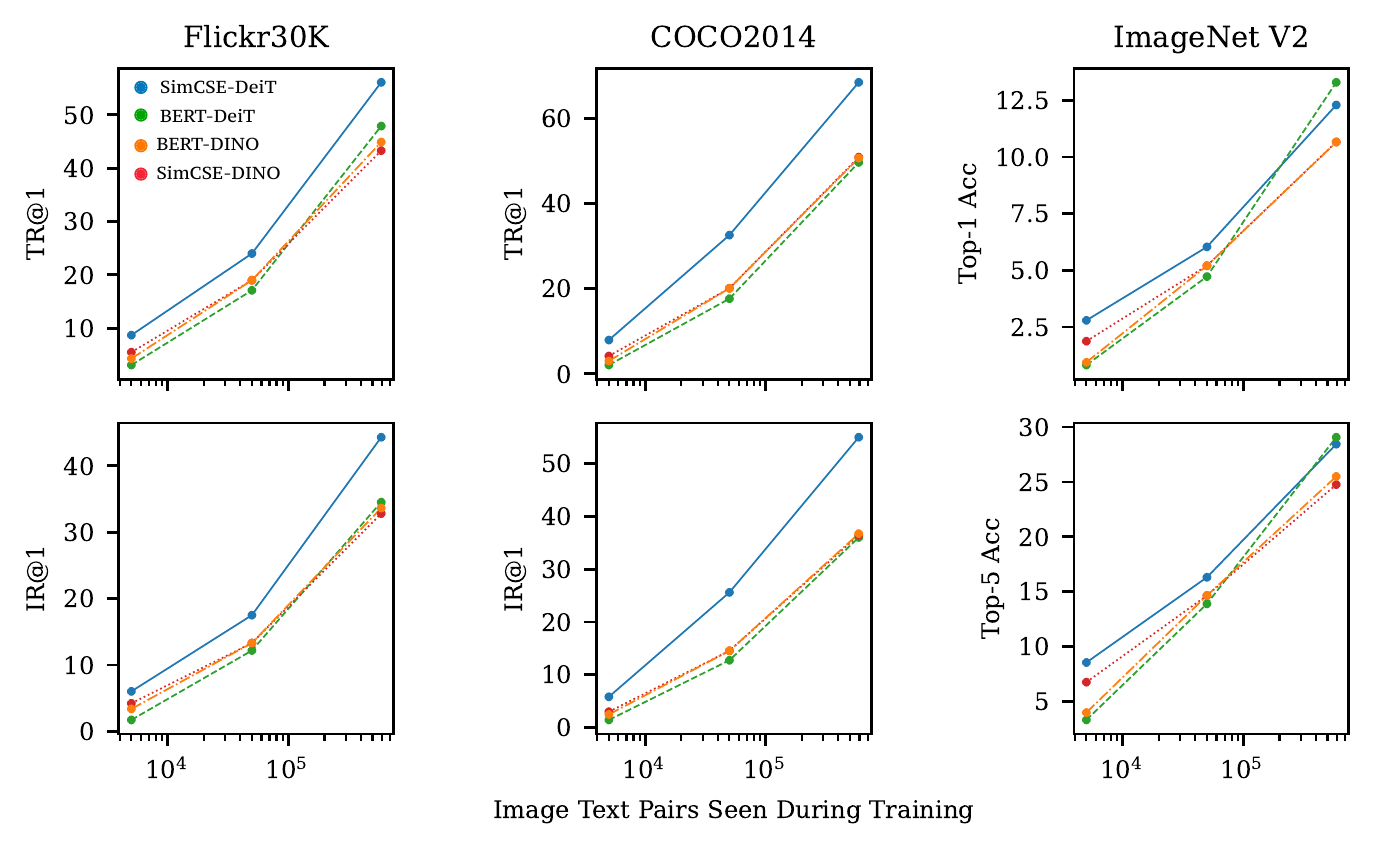}
    \caption{A comparison of different kinds of pretraining on LilT performance. Each model is trained on 591k pairs.}
    \label{fig:pretraining-type}
\end{figure}
We train LilT-base models with encoders initialized from different kinds of pretraining methods.
For the text encoder, we choose between \texttt{bert-base-uncased} \cite{bert} and SimCSE \cite{simcse}.
For the image encoder, we choose between DeiT\cite{deit} and DINO \cite{dino}.
We train all models on 591k pairs following \S\ref{sec:implementation-details}.
The unimodal pretraining methods chosen do have an effect on the performance on the vision-language model.
The combination of SimCSE and DeiT appears to be consistently better than other combinations, although on ImageNetV2, BERT-DeiT performs better.
\subsection{Zero-shot Prompts}
Although CLIP\cite{clip} uses a prompt ensemble, we use only a single prompt for all datasets except SVHN: \texttt{a photo of \{ \}}.
For SVHN, we use the prompt \texttt{a photo of the number \{ \}}.
\subsection{Encoder Symmetry}
\label{sec:encoder-symmetry}
\begin{figure}
  \centering
  \includegraphics{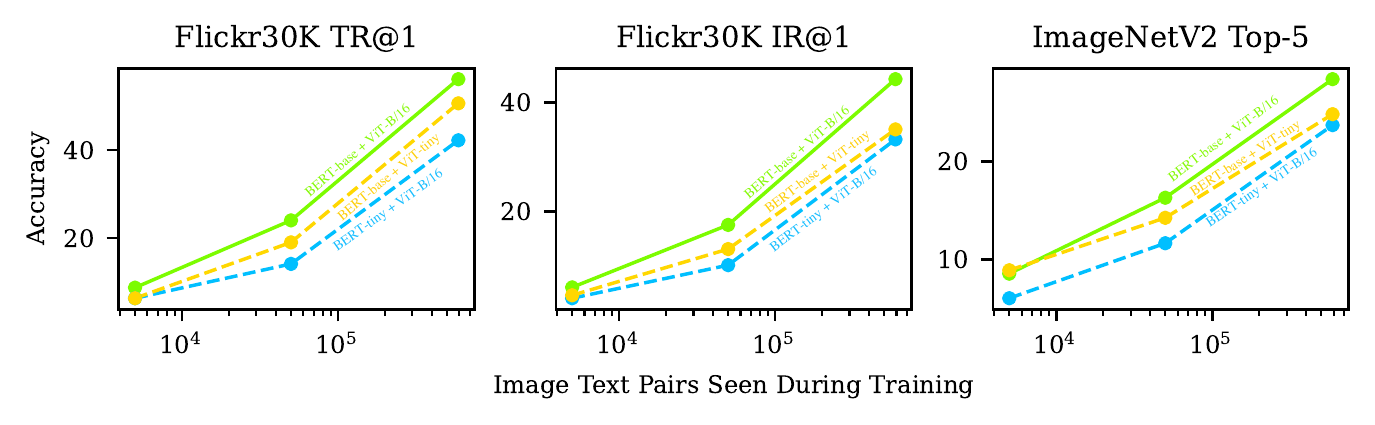}
  \caption{CLIP appears to be more sensitive to the size of the text encoder than the size of the image encoder.}
  \label{fig:encoder-symmetry}
\end{figure}

Which encoder matters more? 
We train three configurations of CLIP on 5k, 50k, 591k pairs (Fig. \ref{fig:encoder-symmetry}).
One is the symmetric CLIP-base, while the two asymmetric configurations have their text encoder and image encoder respectively replaced with the "tiny" version.
Across all three dataset scales, the model with the smaller text encoder performs worse.
\citet{lit} find that on large scale data ($10$m+ pairs), the opposite holds true --- a larger image encoder is better than a larger language model. 
\subsection{Does LilT work with smaller models and less data?}
\begin{figure}
  \centering
  \includegraphics{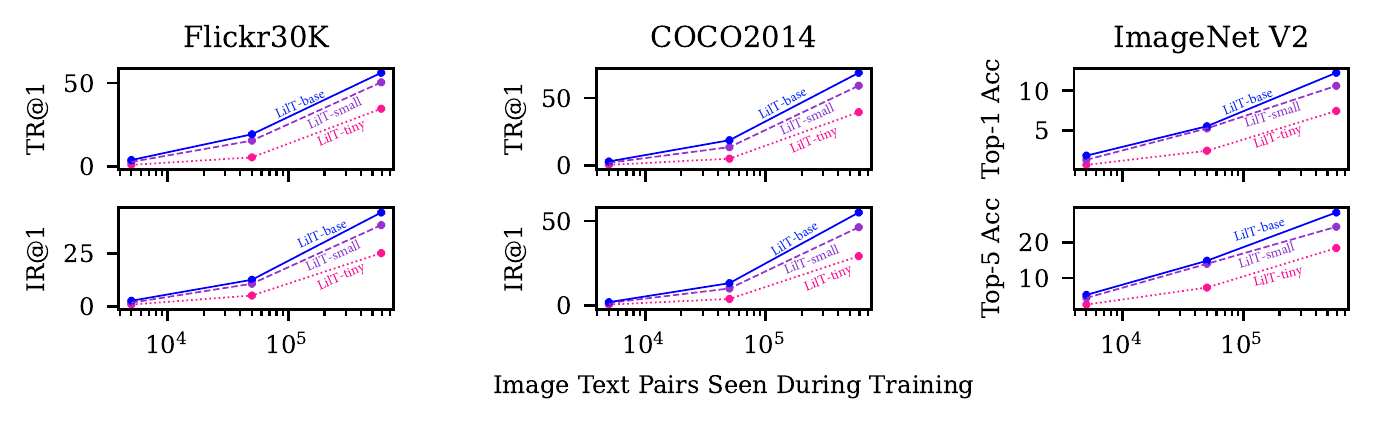}
  \caption{LilT's performance scales with increasing model size and dataset size --- it is not limited to a specific model size or dataset size. LilT$_{DA}$ is pictured.}
  \label{fig:lilt-scaling}
\end{figure}

\begin{figure}
  \centering
  \includegraphics{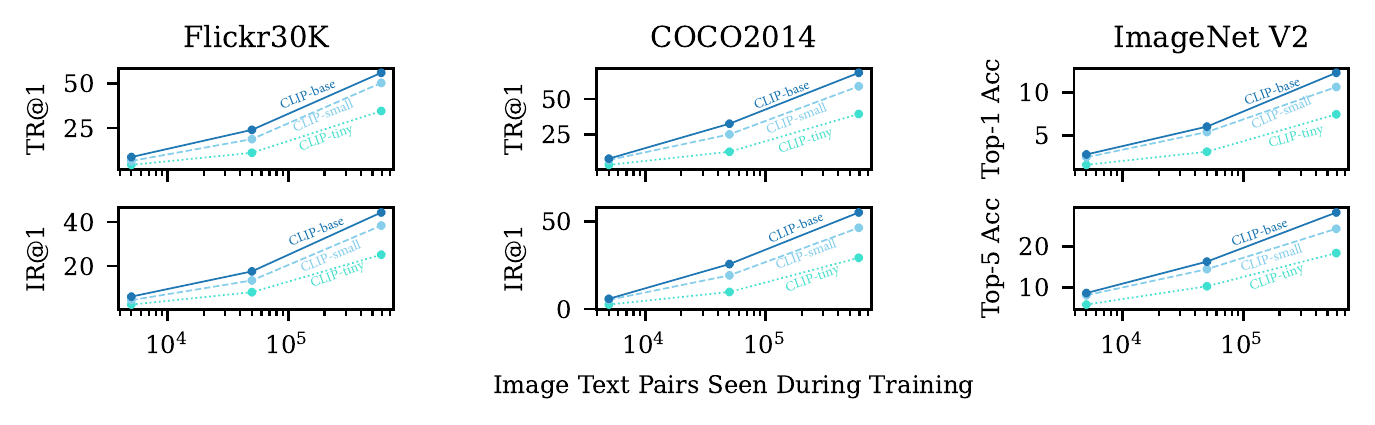}
  \caption{The performance of full-model training on smaller models and with less data.}
  \label{fig:clip-scaling}
\end{figure}

We test LilT and full-model training on smaller versions of transformers, corresponding to `bert-base`, `bert-small`, `bert-tiny`, and with decreasing amounts of image-text pairs (5k, 50k). 
The results are depicted in Figure \ref{fig:lilt-scaling} and Figure \ref{fig:clip-scaling} for LilT$_DA$.
There are no idiosyncratic results --- as model size is decreased, performance decreases for both full model training and parameter efficient transfer learning.
Similarly, as the amount of data decreases, performance also decreases.
This holds true for all tested combinations of dataset size and model size.

\end{document}